# Memory-based Controllers for Efficient Data-driven Control of Soft Robots

Yuzhe Wu[1] and Ehsan Nekouei[2]

## Abstract

Controller design for soft robots is challenging due to nonlinear deformation and high degrees of freedom of flexible material. The data-driven approach is a promising solution to the controller design problem for soft robots. However, the existing data-driven controller design methods for soft robots suffer from two drawbacks: (i) they require excessively long training time, and (ii) they may result in potentially inefficient controllers. This paper addresses these issues by developing two memory-based controllers for soft robots that can be trained in a data-driven fashion: the finite memory controller (FMC) approach and the long short-term memory (LSTM) based approach. An FMC stores the tracking errors at different time instances and computes the actuation signal according to a weighted sum of the stored tracking errors. We develop three reinforcement learning algorithms for computing the optimal weights of an FMC using the Q-learning, soft actor-critic, and deterministic policy gradient (DDPG) methods. An LSTM-based controller is composed of an LSTM network where the inputs of the network are the robot's desired configuration and current configuration. The LSTM network computes the required actuation signal for the soft robot to follow the desired configuration. We study the performance of the proposed approaches in controlling a soft finger where, as benchmarks, we use the existing reinforcement learning (RL) based controllers and proportional-integral-derivative (PID) controllers. Our numerical results show that the training time of the proposed memory-based controllers is significantly shorter than that of the classical RL-based controllers. Moreover, the proposed controllers achieve a smaller tracking error compared with the classical RL algorithms and the PID controller.

**Keywords**-Soft Robots; LSTM; Finite Memory Controllers

[1] Yuzhe Wu is with the City University of Hong Kong (Dongguan). Email: yuzhewu2-c@my.cityu.edu.hk

[2] Ehsan Nekouei is with the department of electrical engineering of City University of Hong Kong. Email: enekouei@cityu.edu.hk

# 1 Introduction

1.1 Motivation

Conventional robotic systems suffer from poor adaptability and flexibility. Soft robots can solve the adaptability and flexibility issues of rigid robots. However, the deformation of soft robots is always nonlinear, and they have infinite degrees of freedom. These facts render the motion planning and control of soft robots a challenging task compared with traditional robotic systems, as the motion of rigid robots can be accurately modeled.

The data-driven control approach offers a promising solution for the control and motion planning of soft robots. In this approach, one can numerically design controllers for a soft robot using finite element simulations or actual measurements. However, data-driven control algorithms for soft robots typically require a large number of samples. Moreover, these algorithms may not achieve the desired performance due to their slow convergence and may result in a locally optimal solution. To highlight this fact, we performed an experiment where we trained three reinforcement learning algorithms, i.e., the soft actor-critic, deep deterministic policy gradient, and deep Q-learning algorithms, to control the position of the tip of a pneumatic soft finger in the SOFA simulation environment. The input to each controller was the position of the tip of the finger, and the training reward was

$$r_t = -1000 \lVert p_t - p_t^d \rVert^2,$$

where $p_t$ and $p_t^d$ are the position of the finger's tip and the desired position of tip at time $t$, respectively.

Fig. 1.1 shows the mean reward of these three algorithms for different episodes, where each episode had 400 samples. We adjusted the layers and the neural network activation functions for each algorithm to achieve the best performance. As Fig. 1.1 shows, even after 250 episodes, these algorithms do not find a desirable controller. This observation suggests that the data-driven control of soft robots can be slow and may result in poorly performing controllers. This observation further motivates us to design fast, data-driven algorithms for designing accurate controllers for soft robots.

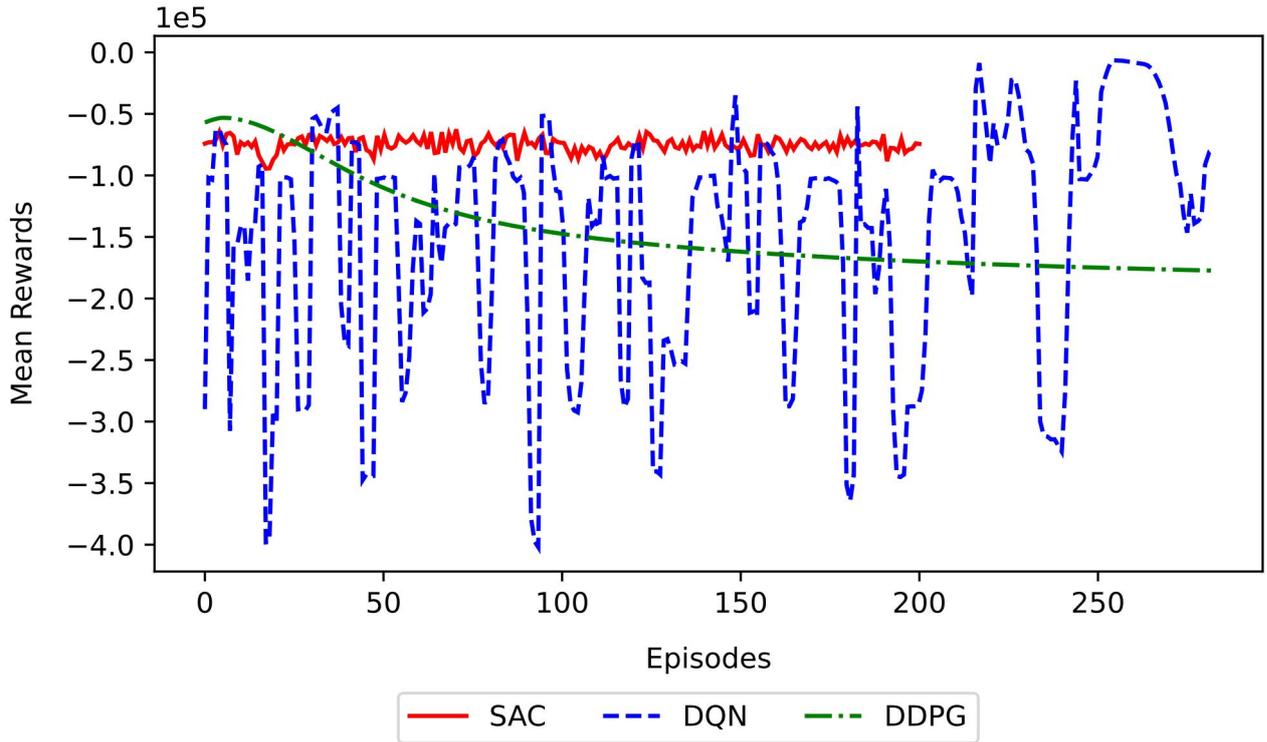

Figure 1.1 Mean rewards per episode of three traditional reinforcement learning algorithms for controlling a soft finger.

1.2 Contributions

In this paper, we develop two novel and efficient data-driven controller design frameworks for soft robots: the LSTM-based framework and the finite memory controller (FMC) framework. In the LSTM approach, an LSTM-based inverse model of a soft robot is developed, which is then used to derive open-loop and closed-loop controllers. Under the FMC framework, the controller input is the current tracking error, i.e., the difference between the current and the desired trajectories. The controller then computes the actuation signal based on the weighted sum of the current and past tracking errors. To compute the optimal weights of an FMC, we propose three different reinforcement learning (RL) algorithms: deep deterministic policy gradient (DDPG), deep Q-network (DQN), and soft actor-critic (SAC) algorithms. Finally, we study the performance of the developed framework for controlling a pneumatically actuated soft finger in the SOFA simulation environment [1] and compare their performance with traditional RL algorithms. Our numerical results show that the developed frameworks outperform the traditional RL algorithms in terms of convergence speed and accuracy.

1.3 Related work

The modeling and control of soft robots have attracted a great deal of attention from academia. Duriez et al. [2] proposed a real-time finite element method (FEM) implementation to model a 3D silicone soft robot actuated by cables accurately. They then used the iterative Gauss-Seidel algorithm to control the robot's position. In [3], Largilliere et al. used a quadratic programming optimization to solve the inverse problem based on the work of Christian Duriez et al. [2] and tested their design in SOFA.

In [4], Bosman et al. modeled a continuum robot with rigid vertebras actuated by cables using the modified FEM and used the Gauss-Seidel algorithm to control the using SOFA. In [5], Katzschmann et al. used reduced-order FEM to model a pneumatically actuated soft robot arm. They controlled the robot moving as desired trajectories using a closed-loop controller with a state observer and tested their work in SOFA. In [6], Bieze et al. used FEM to model a continuum manipulator actuated by both cables and pneumatic actuators. They controlled the robot's tip using a closed-loop controller based on the forward and inverse kinematic models. Their work was implemented in SOFA.

In [7], Hofer and D' Andrea designed a pneumatic soft robotic arm. They designed a closed-loop controller to control the angles of the soft robotic arm so that the robotic arm can keep a rigid link in a vertical position. In [8], Bui et al. modeled the fabric-reinforced inflatable soft robot using discs and cables to simplify the modeling and controller design. They used a linear parameter-varying system to design the controller. In [9], Giorelli et al. used the direct and inverse kinetics models to design a PID controller to control a cable-driven silicone.

Design data-driven controllers have been used for soft robots to avoid complex physical models of soft robots. In [10], Thuruthel et al. designed a trajectory tracking controller for a soft manipulator using recurrent neural networks. In [11], Qiao et al. designed a closed-loop controller to control a pneumatic soft arm's bending and elongation motions using the iterative learning control and a PD feedback controller.

In [12], Xie et al. proposed a data-driven cyclic-motion generation (CMG) scheme with a dynamic neural network for the CMG task of the Sawyer manipulator with unknown models. Centurelli et al. [13] developed a neural network-based closed-loop controller trained by a trust region policy optimization algorithm for a pneumatic actuator. Thuruthel et al. [14] designed a closed-loop force controller using an LSTM network and tested it on a passive anthropomorphic finger.

Goharimanesh et al. [15] proposed a fuzzy reinforcement learning approach for continuum robot control, and their simulation results illustrate a steady and accurate trajectory tracking capability.

Satheeshbabu et al. [16] implemented a model-free reinforcement learning control policy based on deep deterministic policy gradient for pneumatic-driven soft continuum arms. Their results illustrate low errors in path tracking. Truby et al. [17] developed a framework based on recurrent neural networks for learning 3D configuration in a pneumatically actuated soft robot. Their results show good performance in predicting the steady-state configuration of the soft robotic arm.

Tang et al. [18] introduced a learning-based controller for position and force tracking in the bathing task for a pneumatic soft robotic arm. Their results show better performance than PILCO [19]. Zhao et al. [20] proposed a control scheme for a pneumatic-driven continuum robot based on the sequence feature extractor and position decoder neural networks. Their results show a high precision in path following tasks. In [21], Ji et al. introduced a multiagent deep Q-network for trajectory tracking tasks under external loads and implemented it on a cable-driven continuum manipulator. Their results show submillimeter tracking accuracy and high stability under different disturbances. Kovari et. al. [22] presented a control strategy based on the Monte-Carlo tree search and policy gradient (PG) algorithm for pressure tracking in pneumatic actuators. Their simulation results show that the robustness and reliability of their control strategy are better than that of the PG algorithm.

Preiss et al. [23] developed a model based on recurrent neural networks for an extended Kalman filter and a model predictive controller to track the trajectory of a thin, soft cylinder. Their experiments illustrate that the RNN-based controller improves tracking performance compared to an open-loop control scheme. Morimoto et al. [24] designed an ensembled lightweight model-free reinforcement learning network (ELFNet) for trajectory tracking and implemented it on a pneumatic continuum arm model. Their simulation results illustrate that ELFNet performs better than soft actor-critic, proximal policy optimization.

## 3 LSTM-based Trajectory Tracking

The trajectory tracking problem aims to design a control input that allows a robot to follow a predefined reference trajectory, as shown in Fig 3.1. In this figure, $u_t$ is the control input, $p_t$ and $p_t^d$ are the actual configuration and the desired configuration at time $t$, respectively. The trajectory tracking goal can be achieved easily for rigid robots as their dynamic behavior can be accurately modeled. However, since soft robots are made of flexible material, they have infinite degrees of freedom, and their deformation is always nonlinear. At the same time, the behavior of soft robots does not follow from the classical solid mechanics. Thus, compared to rigid robots, the trajectory tracking of soft robots is more complex and difficult.

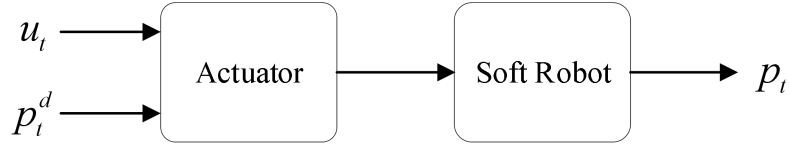

Figure 3.1. The input and output of a soft robot.

In this section, we develop two novel trajectory tracking approaches for soft robots using LSTM neural networks. To this end, we first propose a data-driven inverse model for soft robots based on LSTM neural networks in the next subsection. We then use the inverse model to develop open-loop and closed-loop trajectory tracking controllers for soft robots.

3.1 LSTM-based Inverse Model

In this subsection, we first build an inverse model for a soft robot using an LSTM neural network. We will then use the inverse model to build open-loop and closed-loop controllers for trajectory tracking. An LSTM neural network is composed of a series of LSTM cells. The structure of an LSTM cell is shown in Fig. 3.2, where $\sigma(\cdot)$ is the activation function.

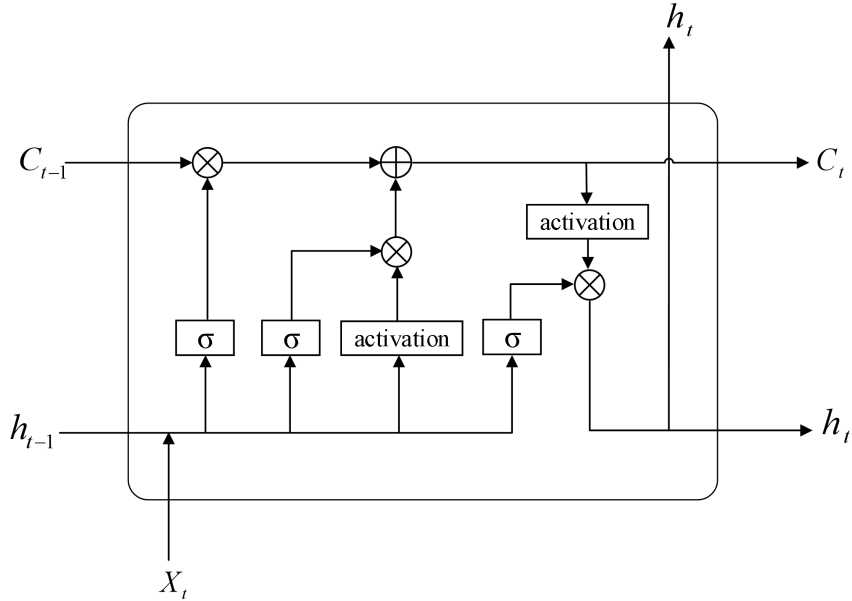

Figure 3.2. The structure of an LSTM cell.

Fig. 3.3 shows an LSTM-based inverse model, which is made of a number of LSTM layers. The input to the inverse model is the current configuration of the robot and its configuration in the next time step, denoted as $p_t$ and $p_{t+1}$, respectively. The output of the inverse model is the required control input, denoted by $u_t$, for the robot to change its configuration from $p_t$ to $p_{t+1}$.

To design an inverse model, we first collect a dataset for training the LSTM network, where we apply a sequence of control inputs to a soft robot and collect the soft robot's corresponding configurations to form the training data. Then, the parameters of the LSTM layers are adjusted such that the difference between the model's output and the control input sequence from the training data is minimized. After training the inverse model, we will use it to perform trajectory tracking.

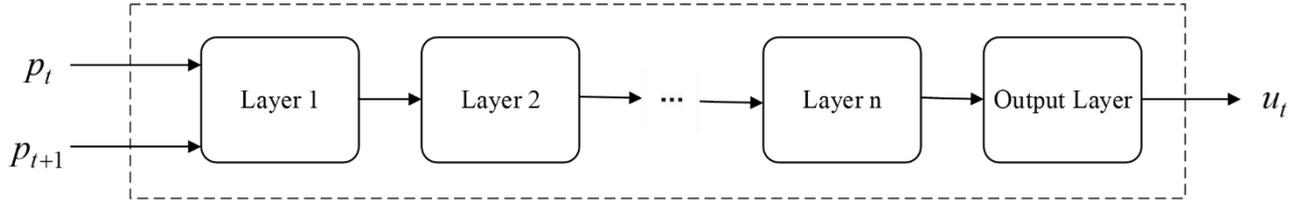

Figure 3.3. The structure of an LSTM-based inverse model.

### 3.1.1. LSTM-based Open-loop Trajectory Tracking

In this subsection, we will use the LSTM-based inverse model to develop an open-loop trajectory tracking controller, as shown in Fig. 3.4. At time $t$, the inputs to the controller are the current desired configuration ($p_t^d$) and the next desired configuration ($p_{t+1}^d$). The controller computes the required actuation signal for moving the soft robot from the current configuration to its next configuration. The open-loop controller provides a simple solution for trajectory tracking as it does not require sensing the position of the soft robot at each time step. However, it is not able to compensate for disturbances and model uncertainties.

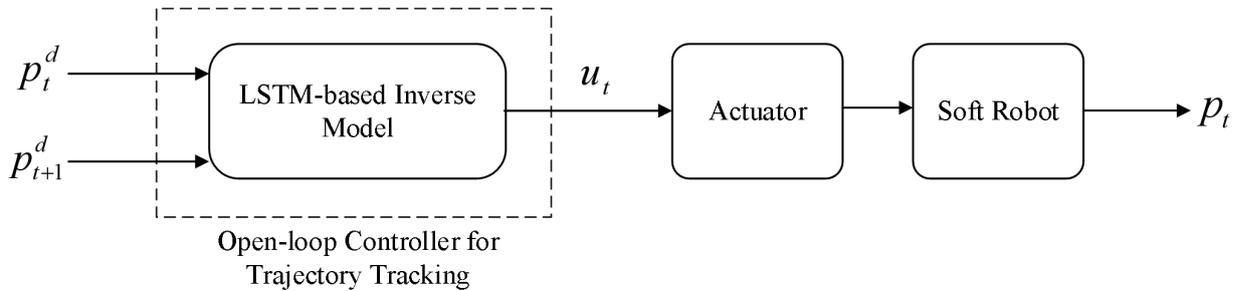

Figure 3.4. The open loop LSTM-based trajectory tracking.

### 3.1.2 LSTM-based Closed-loop Trajectory Tracking

In this subsection, we will construct a closed-loop trajectory tracking controller using the LSTM-based inverse model, as shown in Fig. 3.5. The inputs to the controller are the current configuration of the soft robot measured by a sensor and its next desired configuration. Using these inputs, the controller computes the actuation signal required for moving the robot from its current configuration ($p_t$) to the desired configuration at time $t + 1$ ($p_{t+1}^d$). Although the closed-loop trajectory tracking controller requires a sensor to measure the configuration, our numerical results show that the closed-loop controller results in a smaller tracking error than the open-loop controller.

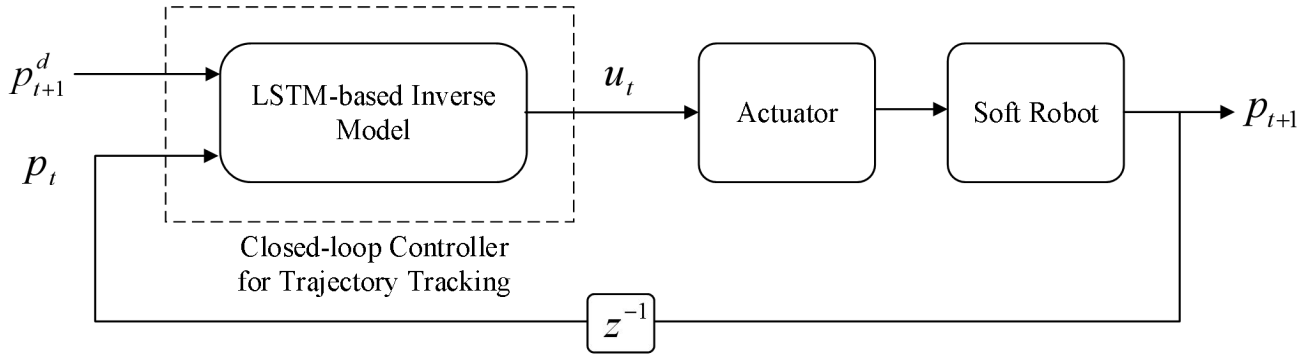

Figure 3.5. The closed-loop LSTM-based trajectory tracking controller.

# 4 Trajectory Tracking Using Finite Memory Controllers

In this section, we develop a trajectory tracking framework for soft robots based on finite memory controllers (FMCs). Fig. 4.1. shows the structure of an FMC with $k$ memory units. At time $t$, the input to the controller is the tracking error, i.e., $e_t = p_t - p_t^d$. The controller uses the current error and the past $k$ errors, which are stored in its memory, to compute the required inputs of actuators to track the desired trajectory, as shown in Fig. 4.2. The output of an FMC is the weighted sum of the current and past errors. The weights of the controller are computed by solving the following optimal control problem

$$\min_{\{w_0, w_1, \cdots, w_k\}} \sum_{t=0}^{N} E\left[\|e_t\|^2\right]$$

where, $e_t$ is the tracking error at time $t$.

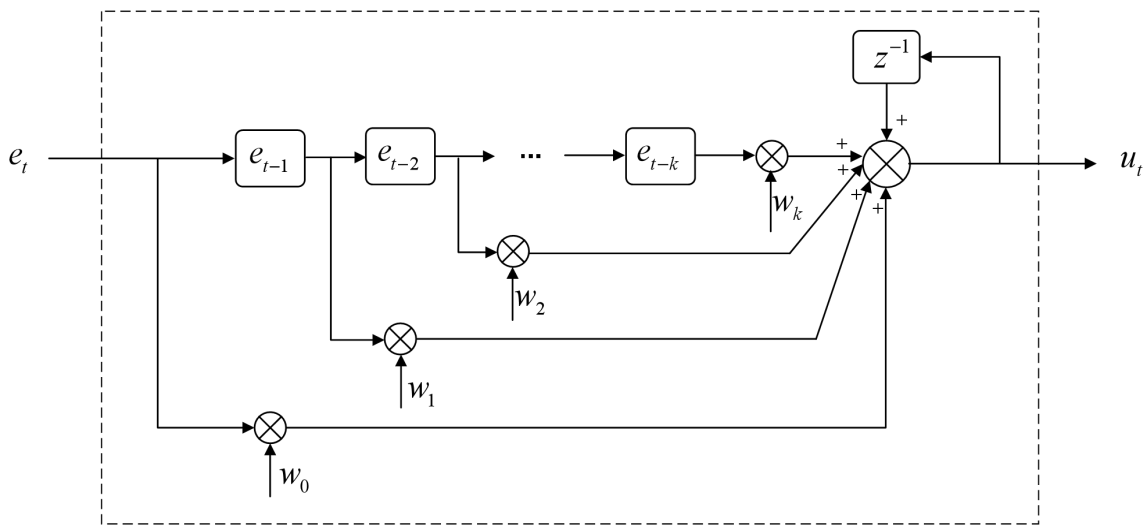

Figure 4.1. The structure of the finite memory controller.

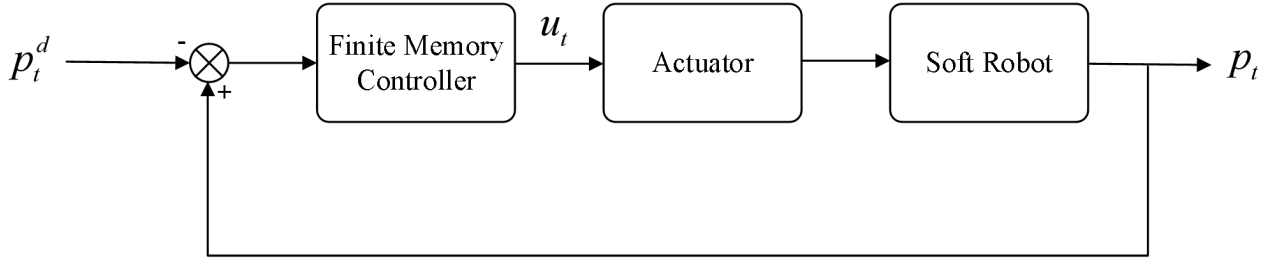

Figure 4.2. The finite memory controller for a soft robot.

We next propose three algorithms for computing the optimal weights of an FMC using deep deterministic policy gradient (DDPG), deep Q-network (DQN), and soft actor-critic (SAC) algorithms.

4.1 FMC Design using DDPG

The DDPG algorithm consists of an actor neural network, a critic neural network, a target actor neural network, a target critic neural network, and a replay buffer. In this algorithm, the state of the controller is defined as $s_t = [e_t, e_{t-1}, e_{t-2}, \cdots, e_{t-k}]$, and the reward function is defined as $r_t = -1000\|p_t - p_t^d\|^2$.

---
**Algorithms 1** FMC with DDPG
---

**Initialization**

Randomly initialize critic network $Q(s, a|w^Q)$ and actor-network $\mu(s|w^\mu)$ with weights $w^Q$ and $w^\mu$.

    Initialize target networks $\bar{Q}$ and $\bar{\mu}$ with weights $w^{\bar{Q}} \leftarrow w^Q$, $w^{\bar{\mu}} \leftarrow w^\mu$.

    Initialize replay buffer $B \leftarrow \emptyset$ and the weights of FMC $w \leftarrow \emptyset$.

    Set the hyperparameters $\gamma, \tau$.

    Initialize the input of actuators as $u_t = 0$.

**for** each episode **do**

    Initialize the configuration of the robot and reward $R_n = 0$ for each episode (*n* is episode index).

        **for** each time step **do**

            Select action $a_t = \mu(s_t|w^\mu)$ according to the current policy.

            Update the input of soft robots $u_t \leftarrow u_t + a_t$.

    Observe reward $r_t$ and obtain new state $s_{t+1}$.

   Store transition $(s_t, a_t, r_t, s_{t+1})$ in $B$.

   Update $R_n \leftarrow R_n + r_t$.

  **for** each gradient step **do**

   Sample a random mini-batch of $N$ transitions $(s_t, a_t, r_t, s_{t+1})$ from $B$.

   Set $y_i = r_i + \gamma \bar{Q}\left(s_{i+1}, (\bar{\mu}(s_{i+1}|w^{\bar{\mu}}))\big|w^{\bar{Q}}\right)$.

   Update the critic network by minimizing the loss: $L = \frac{1}{N}\sum_i \left(y_i - Q(s_i, a_i|w^Q)\right)^2$.

   Update the actor policy using the sampled policy gradient:

$$\nabla_{w^\mu} J \approx \frac{1}{N}\sum_i \nabla_a Q(s, a|w^Q)\big|_{s=s_i, a=\mu(s_i)} \nabla_{w^\mu}\mu(s|w^\mu)\big|_{s_i}.$$

   Update the target networks:

$$w^{\bar{Q}} \leftarrow \tau w^Q + (1-\tau)w^{\bar{Q}}$$
$$w^{\bar{\mu}} \leftarrow \tau w^\mu + (1-\tau)w^{\bar{\mu}}.$$

  Set the mean episode reward as $\overline{R_n} = R_n/(number\ of\ time\ steps)$

  **if** $\overline{R_n} > \overline{R_{n-1}}$ **then**

   Update the weights of the FMC as $w \leftarrow w^\mu$.

**Output**

 Actor $\mu(s|w)$ and weights $w$.

## 4.2 FMC Design using DQN

  The DQN consists of a Q-network and a target Q-network. The weights of these networks are updated according to Algorithm 2.

**Algorithms 2** FMC with DQN

**Initialization**

Randomly initialize the Q-network with weights $w^Q$.

Initialize target Q-network as $w^{\bar{Q}} \leftarrow w^Q$.

Initialize replay buffer as $B \leftarrow \emptyset$ and the weights of FMC as $w \leftarrow \emptyset$.

 Initialize the action space of the robot as $F \leftarrow [f_0,\ f_1,\ \cdots,\ f_m]$.

 Set the discount factor $\gamma$.

 **for** each episode **do**

  Initialize the configuration of the robot and rewards $R_n = 0$ for each episode ($n$ is episode index).

    **for** each time step **do**

      With probability $\epsilon$ select a random action $a_t$.

      Otherwise select action $a_t = argmax_a Q(s_t, a; w^Q), a \in F$ according to the current policy.

        Update the input of soft robots $u_t \leftarrow u_t + f_{a_t}$.

        Observe reward $r_t$ and obtain new state $s_{t+1}$.

        Store transition $(s_t, a_t, r_t, s_{t+1})$ in $B$.

        Update $R_n \leftarrow R_n + r_t$.

        Every $C$ steps reset $w^{\bar{Q}} \leftarrow w^Q$.

    **for** each gradient step **do**

      Sample a random mini-batch of $N$ transitions $(s_t, a_t, r_t, s_{t+1})$ from $B$.

      Set $y_i = \begin{cases} r_i & \text{if episode terminates at step } i+1 \\ r_i + \gamma max_{\bar{a}} \bar{Q}(s_{i+1}, \bar{a}; w^{\bar{Q}}) & \text{otherwise} \end{cases}$

      Perform a gradient descent step on $\left(y_i - Q(s_i, a_i | w^Q)\right)^2$ with respect to $w^Q$.

      Set the mean episode reward as $\overline{R_n} = R_n / (number\ of\ time\ steps)$.

      **if** $\overline{R_n} > \overline{R_{n-1}}$ **then**

        Update $w \leftarrow w^Q$

**Output**

    Actor $\mu(s|w)$ and weights $w$.

## 4.3 FMC Design using SAC

The SAC consists of a Q-value critic network and an actor network where their weights are updated according to Algorithm 3.

**Algorithms 3** FMC with SAC

**Initialization**

    Randomly initialize the weights of the Q-value critic network and the actor network $A$.

    Initialize V critic network $V$ and target V critic network $\bar{V}$ with weights $\psi, \bar{\psi}$.

    Initialize replay buffer $B \leftarrow \emptyset$ and the weights of FMC $w \leftarrow \emptyset$.

    Initialize the input of actuators of soft robots $u_t = 0$.

**for** each episode **do**

    Initialize the position of the robot and rewards $R_n = 0$ for each episode ($n$ is episode index).

    **for** each time step **do**

Select action $a_t = \pi_\phi(\cdot|s_t)$ according to the current policy.

Update $u_t \leftarrow u_t + a_t$

Observe reward $r_t$ and obtain new state.

Store transition $(s_t, a_t, r_t, s_{t+1})$ in $B$.

Update $R_n \leftarrow R_n + r_t$.

**for** each gradient step **do**

Sample a random mini-batch of $N$ transitions $(s_t, a_t, r_t, s_{t+1})$ from $B$.

Update the V critic network, the target V critic network and the Q networks as in [27].

Set the mean episode reward as $\overline{R_n} = R_n/(number\ of\ time\ steps)$

**if** $\overline{R_n} > \overline{R_{n-1}}$ **then**

Update $w \leftarrow \phi$.

**Output**

Actor $A$ and weights $w$.

## 5 Simulation results

In this section, we study the performance of the proposed trajectory tracking frameworks in controlling the tip of a pneumatic soft finger. To this end, we modeled a pneumatic soft finger using the SOFA simulation framework, as shown in Fig. 5.1. In our simulation environment, the soft finger can only move in a plane, i.e., a two-dimensional space. Thus, to simplify the controller design, we use the angle between the tip and the base of the finger, denoted as $\theta_k$ in Fig. 5.1, to specify the position of the soft finger's tip at time step $k$. The objective of a trajectory tracking controller is to ensure that the tip of the finger follows a desired reference trajectory denoted by $\theta^r = [\theta_0^r, ..., \theta_k^r]$, where $\theta_k^r$ is the reference point at time $k$.

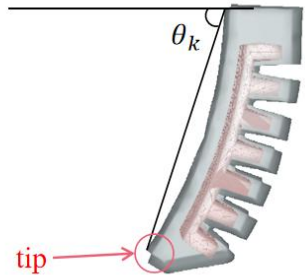

Figure 5.1. A pneumatic-driven soft finger.

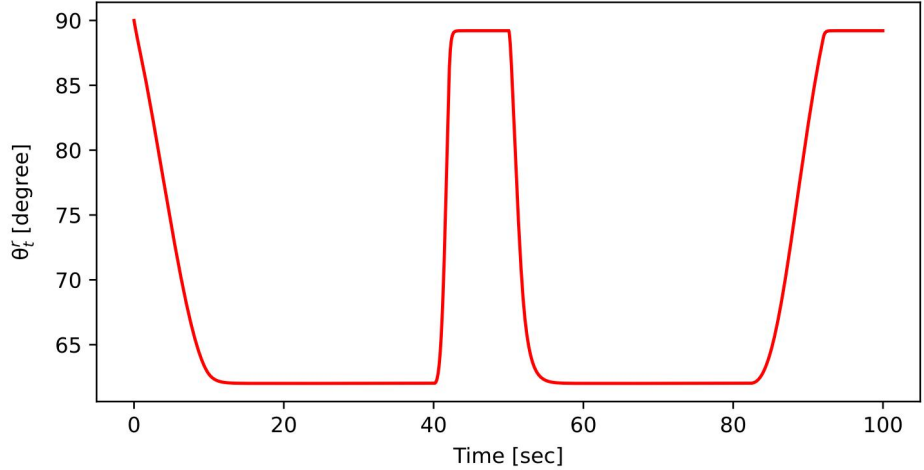

Figure 5.2. The reference trajectory of the soft robot.

Fig. 5.2. shows the reference trajectory used in our simulations. In the rest of the section, we will study the tracking performance of the LSTM-based and finite memory controllers.

5.1 Open-loop LSTM-based Controller

To build the LSTM model for trajectory tracking, we applied a pressure trajectory to the soft finger in SOFA and recorded the tip's position. Then, we used the recorded data to build an LSTM-based inverse model, as shown in Fig. 3.4. We then applied the reference trajectory to the inverse model to obtain the pressure sequence. The training trajectory had different shapes and frequencies from the test trajectory that was used to evaluate the controller's performance.

Fig. 5.3. shows the tracking results of the open-loop LSTM-based controller, and Fig. 5.4 shows its corresponding tracking error. According to Fig. 5.3. and Fig. 5.4, the open-loop LSTM-based controller cannot accurately track the reference trajectory when the reference trajectory changes rapidly. Moreover, the steady-state tracking error of the controller is relatively large.

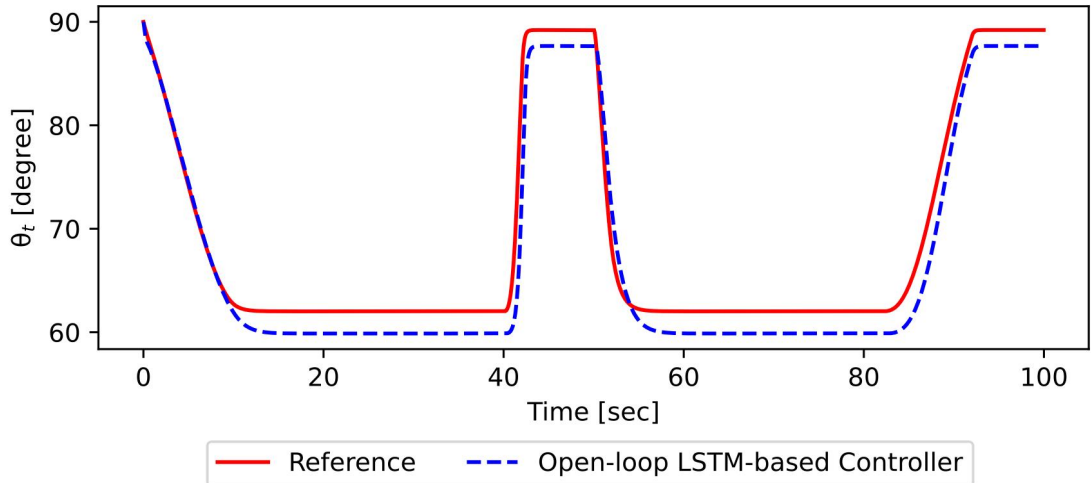

Figure 5.3. Tracking results of the open loop LSTM model.

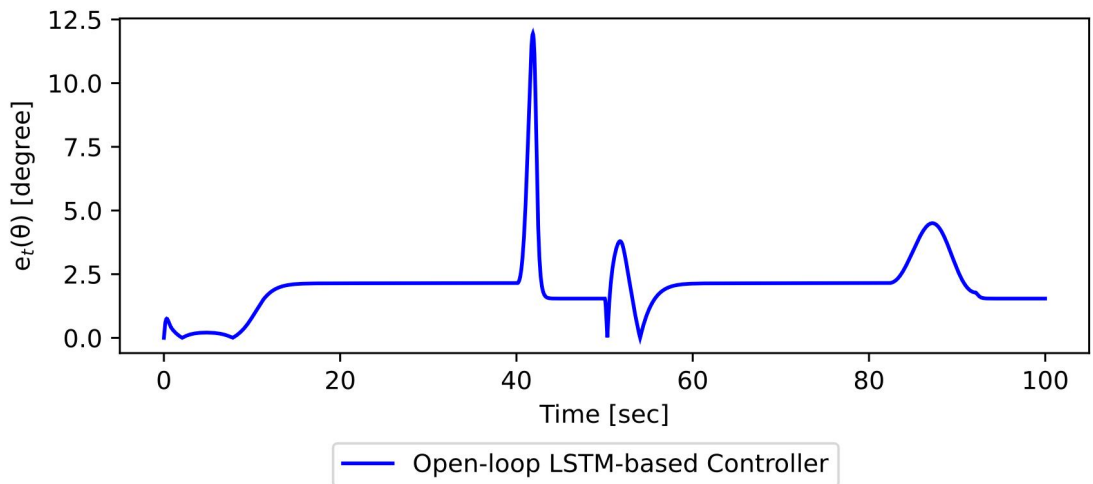

Figure 5.4. Tracking errors of the open loop LSTM model.

5.2 Closed-loop LSTM-based trajectory tracking

To build the closed-loop LSTM-based controller, we used the same method as in subsection 5.1 to train the LSTM model. Fig. 5.5. shows the tracking results of the closed-loop LSTM-based controller. According to this figure, the controller can closely track the reference trajectory. At this point, it is constructive to compare the performance of the closed-loop LSTM-based controller with that of a proportional-integral (PI) controller. Fig. 5.6. shows the tracking error of the closed-loop LSTM-based controller and that of the PI controller. We adjusted the parameters of the PI controller to achieve the best tracking performance. According to this figure, the closed-loop LSTM-based controller can closely track the reference trajectory even when the reference trajectory changes rapidly. Moreover, the steady-state error of the LSTM-based controller is close to zero. Finally, according to Fig. 5.6, the tracking error of the LSTM-based controller is less than that of the PI controller. Based on Fig. 5.5. and Fig. 5.6, the closed-loop LSTM model shows a desirable performance in trajectory tracking compared with the open-loop LSTM model and the PI controller.

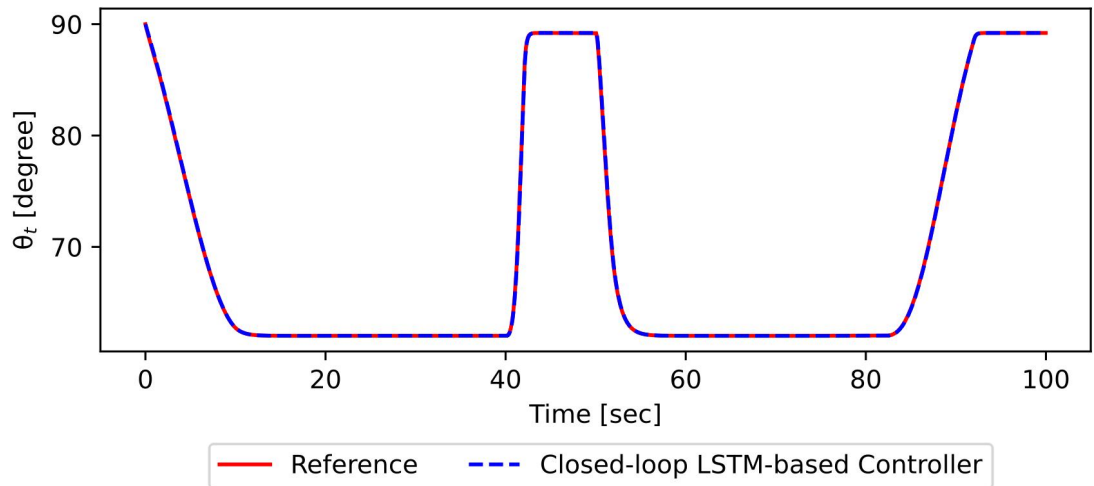

Figure 5.5. Tracking results of the closed-loop LSTM-based controller.

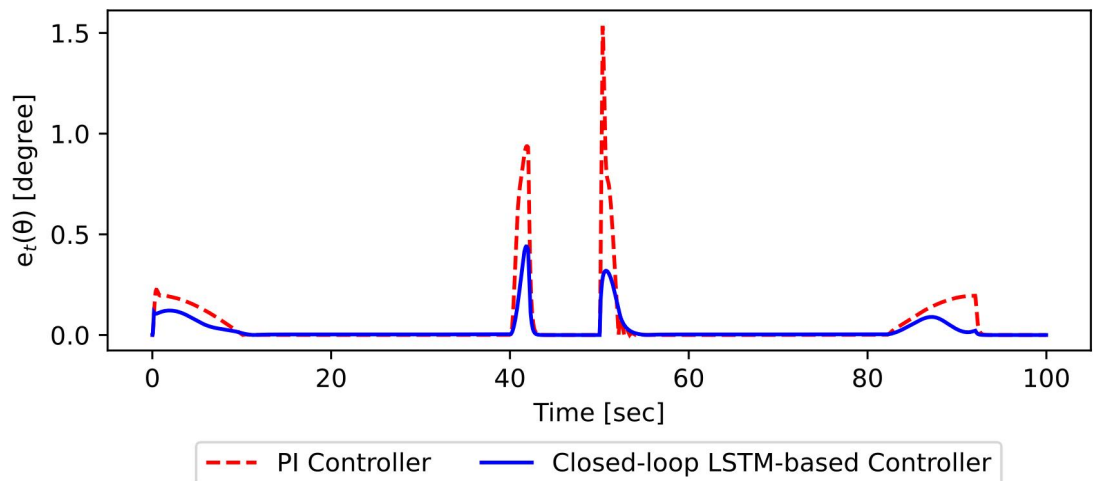

Figure 5.6. Tracking errors of the closed loop LSTM model and the PI controller.

5.3 Finite Memory Controller

To build the finite memory controller (FMC), we used SOFA to compute the best weights of an FMC with four memory units based on the DDPG, DQN, and SAC algorithms. After computing the optimal weights, we used a test reference trajectory to study the tracking performance of the FMC in SOFA. The test trajectory had different shapes and frequencies from the training trajectory. For the FMC with DDPG, we used a hyperbolic tangent as the activation function to ensure the output of DDPG would be constrained to a suitable interval. For the FMC with DQN, we used a Sigmoid function as the activation function. For the FMC with SAC, we used a linear function as the activation function to obtain a proper action value. We used a single-layer actor neural network for all these controllers to avoid a large training time.

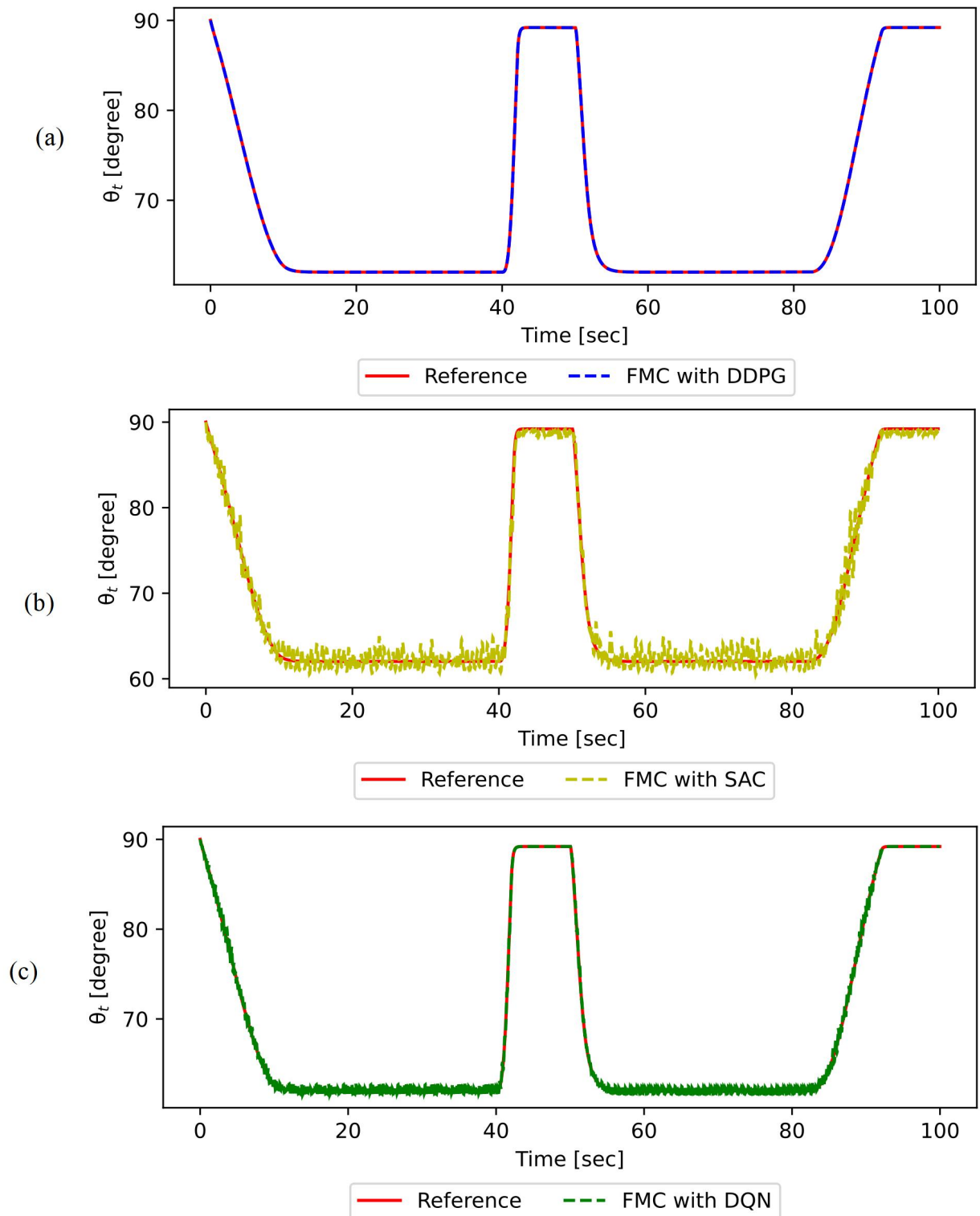

Figure 5.7. Tracking results of the finite memory controllers with different RL algorithms.

Fig. 5.7 shows the performance of the FMC with DDPG, DQN, and SAC algorithms. According to this figure, the FMC with DDPG can accurately track the reference trajectory. Moreover, the FMCs with DQN and SAC are capable of tracking the reference trajectory, but these controllers result in fluctuations around the reference trajectory due to the stochastic nature of these algorithms.

The tracking error of the FMC with different RL algorithms is shown in Fig. 5.8. According to this figure, the maximum error of the FMC with DDPG is 0.04 degrees, whereas the maximum tracking errors of FMC with DQN and SAC are around 1.5 degrees and 5 degrees, respectively. This observation confirms that the FMC with DDPG achieves the best tracking performance.

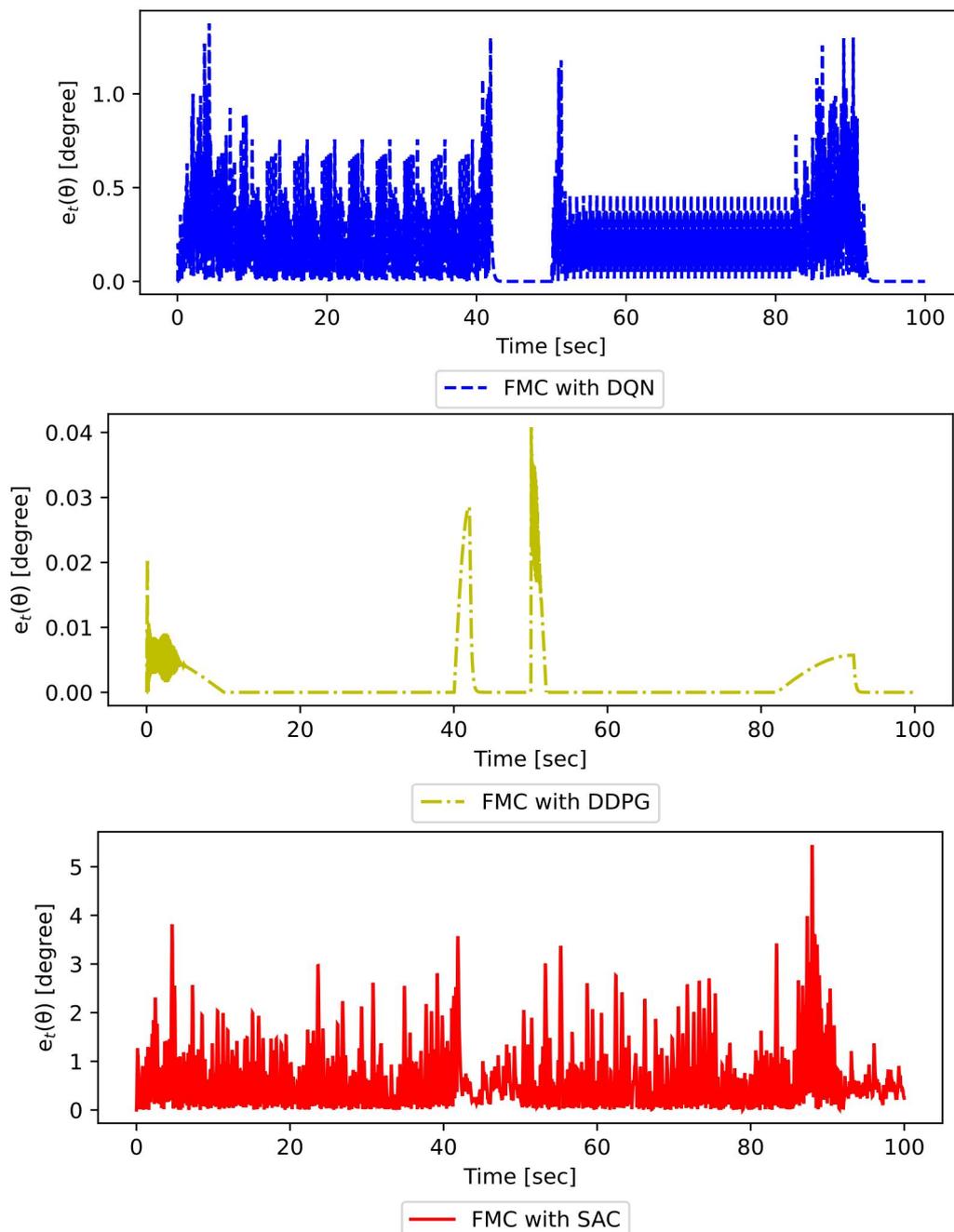

Figure 5.8. Tracking errors of the finite memory controllers with different RL algorithms.

We next compare the training performance of the FMCs with that of the traditional RL algorithms. Fig. 5.9. shows the mean reward per training episode for FMCs (trained with DDPG, DQN, and SAC algorithms) and the mean reward of standard neural network controllers (trained by DQN, SAC, and DDPG algorithms). To obtain the best performance, we adjusted the layers of the neural networks and their activation functions for each algorithm. For all the RL algorithms, we used three layers for the actor-network. For the DDPG algorithm, we used the Relu function as the activation function. For the DQN algorithm, we used the Sigmoid function as the activation function; for the SAC algorithm, we used the Relu function as the activation function.

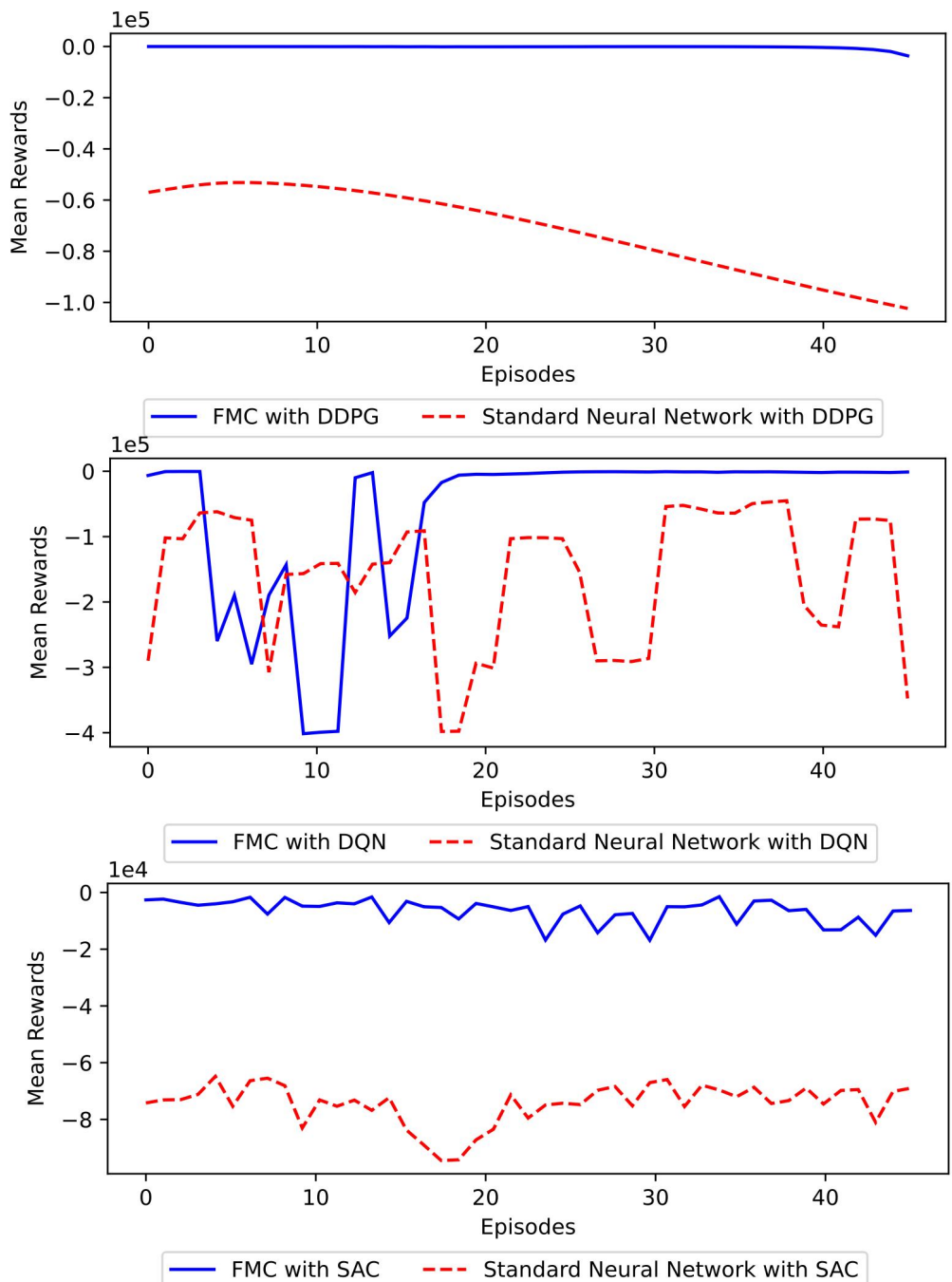

Figure 5.9. The mean reward of the finite memory controllers and the traditional RL algorithms.

We finally compare the performance of the closed-loop LSTM-based controller, an FMC trained with DDPG, and a PI controller with optimal gains in Fig. 5.10. According to this figure, the FMC with DDPG achieves the best tracking performance with a tracking error close to zero, where is the maximum tracking errors of the closed-loop LSTM and the PI controllers are 0.5 degrees and 1.5 degrees, respectively.

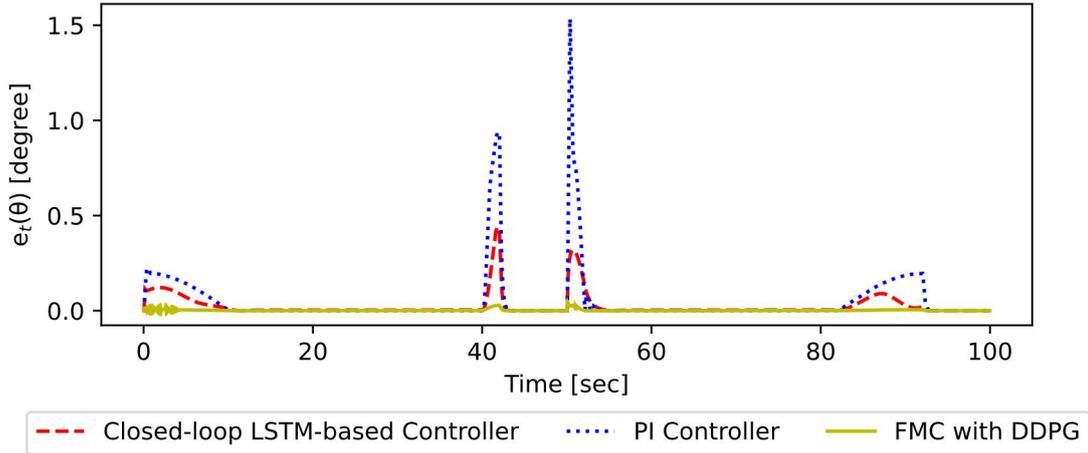

Figure 5.10. Tracking errors of the closed-loop LSTM, FMC, and the PI controller.

Table 5.1 summarizes the average tracking error of different controllers. The average tracking error is defined as $e(\theta) = \frac{1}{N+1}\sum_{t=0}^{t=N}|e_t(\theta)|$ where, $e_t(\theta) = \theta_t - \theta_t^r$ is the error at time $t$. According to Table 5.1, the FMC with the DDPG algorithm achieves the smallest average tracking error, followed by the closed-loop LSTM-based controller. Besides, the finite memory controller used much less training data (only 400 samples), while the closed-loop LSTM model used 20,000 samples for training.

Table 5.1. The average trajectory tracking errors of different controllers

| Controller | | Average Trajectory Tracking Error |
|---|---|---|
| PI Controller | | 0.05369 |
| Open-loop LSTM-based | | 2.02744 |
| Closed-loop LSTM-based | | 0.02454 |
| FMC | SAC Algorithm | 0.75345 |
| | DQN Algorithm | 0.25283 |
| | DDPG Algorithm | 0.00154 |

## Conclusion

In this paper, we developed LSTM-based inverse models and finite memory controllers for trajectory tracking for soft robots. We also studied the performance of the developed controllers using a pneumatic-driven soft robot in the SOFA simulation environment and compared the performance of the developed controllers with that of traditional RL-based controllers and the PI controller. According to our numerical results, the closed-loop LSTM-based inverse model and FMC with DDPG achieve the best performance in trajectory tracking.